\documentclass[fleqn,10pt]{wlscirep}
\usepackage[utf8]{inputenc}
\usepackage[T1]{fontenc}
\usepackage{textcomp, gensymb}
\usepackage[parfill]{parskip}
\usepackage{amsmath}
\usepackage[inkscapelatex=false]{svg}
\usepackage{subcaption}
\usepackage{placeins}

\title{Three-Dimensional, Multimodal Synchrotron Data for Machine Learning Applications}

\author[1,2,*]{Calum Green}
\author[2]{Sharif Ahmed}
\author[2]{Shashidhara Marathe}
\author[2]{Liam Perera}
\author[2]{Alberto Leonardi}
\author[1]{Killian Gmyrek}
\author[1]{Daniele Dini}
\author[3,4]{James Le Houx}

\affil[1]{Imperial College London, Department of Mechanical Engineering, London, SW7 2AZ, United Kingdom}
\affil[2]{Diamond Light Source, Rutherford Appleton Laboratory, Didcot, OX11 0QX, United Kingdom}
\affil[3]{ISIS Neutron \& Muon Source, Rutherford Appleton Laboratory, Didcot, OX11 0QX, United Kingdom}
\affil[4]{The Faraday Institution, Harwell Science and Innovation Campus, Didcot, OX11 0RA, United Kingdom}

\affil[*]{corresponding author: Calum Green (cg1417@ic.ac.uk)}

\begin{abstract}

Machine learning techniques are being increasingly applied in medical and physical sciences across a variety of imaging modalities; however, an important issue when developing these tools is the availability of good quality training data. Here we present a unique, multimodal synchrotron dataset of a bespoke zinc-doped Zeolite 13X sample that can be used to develop advanced deep learning and data fusion pipelines. Multi-resolution micro X-ray computed tomography was performed on a zinc-doped Zeolite 13X fragment to characterise its pores and features, before spatially resolved X-ray diffraction computed tomography was carried out to characterise the homogeneous distribution of sodium and zinc phases. Zinc absorption was controlled to create a simple, spatially isolated, two-phase material. Both raw and processed data is available as a series of Zenodo entries. Altogether we present a spatially resolved, three-dimensional, multimodal, multi-resolution dataset that can be used for the development of machine learning techniques. Such techniques include development of super-resolution, multimodal data fusion, and 3D reconstruction algorithm development.

\end{abstract}

\begin{document}

\flushbottom
\maketitle

\thispagestyle{empty}

\section*{Background \& Summary}


Machine learning and the use of deep learning architectures such as convolutional neural networks (CNNs), generative adversarial networks (GANs) and diffusion models have successfully been adopted in the medical imaging community for tasks such as segmentation, classification, super-resolution, and data fusion. \cite{wu_medsegdiff-v2_2023, p_brain_2022, gu_medsrgan_2020, perez-raya_towards_2020} Data fusion, the combination of two or more datasets (modalities), is a powerful tool that can create a new \textit{fused} dataset that contains the best features of each dataset such as resolution and field-of-view, or contains complementary information such as chemical and spatial information. \cite{van_de_plas_image_2015} Deep learning techniques are implemented on 3D X-ray imaging within the physical sciences, such as the digital rocks community using GANs for super-resolution sandstone images; Li-ion battery research using 3D U-Nets for electrode segmentation of 3D volumes; and even fusion of complementary 2D and 3D imaging datasets of different spatial resolutions. \cite{shan_rock_2022, muller_deep_2021,dahari_fusion_2023} However, a key issue when working with deep learning architectures is the availability of good quality training data. For super-resolution tasks, this presents itself in the form of having spatially aligned high and low-resolution images. Whereas for data fusion tasks availability of two spatially resolved modalities of the same sample is extremely important, but rarely exists or is difficult to obtain.

Synchrotron facilities provide extremely large photon flux, and can provide higher spatial and temporal resolutions than traditional lab-based X-ray equipment and experiments. They also feature the capability of having an experimental setup that can allow simultaneous acquisition of different modalities in-situ. Such setups include the capability to obtain imaging and diffraction data to characterise spatial information and understand the phase description as is performed on the K11 beamline at Diamond Light Source. 

In this work present a multi-resolution X-ray Computed Tomography (XCT) dataset acquired on the I13-2 micro-imaging beamline, and a spatially resolved X-ray Diffraction Computed Tomography (XRD-CT) dataset from the K11 Dual Imaging and Diffraction beamline (DIAD) of a bespoke, partially zinc-doped zeolite 13X particle (Zn-13X). The key aim of this dataset is to provide spatially correlated, multi-resolution, multimodal data that is large enough to sufficiently train deep learning models.

\subsection*{Zinc-Doped Zeolite 13X}

Zeolite 13X (13X) is an aluminosilicate material ($Na_{86}[(AlO_{2})_{86}(SiO_{2})_{106}] \cdot xH_{2}O$) used commonly in the carbon capture industry due its ability to store carbon dioxide within the pores of its crystalline framework. It features a nano-porosity with an average nano-pore diameter of $9$\r{A}, and features larger macro and mesopores.\cite{su_co2_2012} These larger pores can be visualised using micro-XCT. 13X also has the ability to undergo ion-exchange to replace the sodium-ions present in the framework.\cite{gal_ion-exchange_1975} Alternative ions include zinc ($Zn^{2+}$), which was chosen as the dopant for this experiment in part to its non-toxic properties. The role of the zinc is to introduce a spatially confined, phase based heterogeneity to create a simple two-phase material that can be characterised using X-ray Diffraction (XRD). Zinc also introduces a k-edge at $9.7keV$ to allow for further modalities such as X-ray Fluorescence Spectroscopy (XRF) to be collected in the future.

\subsection*{Datasets and Reuse Potential}

We present two complimentary X-ray Imaging datasets acquired on the DIAD and I13-2 beamlines. Multi-resolution XCT was acquired at four pixel-sizes: $2.6\mu m$, $1.625\mu m$, $0.8125\mu m$ and $0.325\mu m$ at field-of-views (FOV) of $6.7 \times 5.6mm$, $4.2 \times 3.5mm$, $2.1 \times 1.8mm$ and $0.83 \times 0.7mm$ respectively on I13-2. The whole sample is in the field-of-view for the three larger pixel-sizes. XCT data at $0.54\mu m$, and $1.7 \times 1.7mm$ FOV was also acquired on DIAD alongside $50\mu m$ pixel-size XRD-CT at a $1 \times 1mm$ FOV, and three region-of-interest slices at $25\mu m$ pixel-size and $1 x 1mm$ FOV. The whole sample is in the field-of-view for all the data acquired on DIAD.

Altogether, seven different datasets are made publicly available in this work, with all data hosted across a series of Zenodo entries. Further detailed information is described in the \textit{Data Records} and \textit{Usage Notes} sections. All data is presented in both raw and processed formats to maximise its use to the wider community.

We propose that the I13-2 XCT dataset be used for benchmarking super-resolution methodologies on experimentally acquired data, rather than having to use synthetic datasets or upsample/downsample an image to create the necessary high/low-resolution image pairs. We also anticipate that the spatially resolved XRD-CT data can be used to develop data fusion methodologies, through combining spatially-resolved datasets of complementary phase and spatial  information that are of different resolutions. All data is provided in both raw and processed forms, and as such will be useful for the development of reconstruction and post-processing algorithms for XCT and XRD-CT.

Figure 1 below is a summary of the experimental setup, showing the sample on the beamline, a single tomographic projection, and a summary of the multimodal data that was acquired for a specific subslice of the zeolite sample. All the modes have been aligned and registered in the figure.

\begin{figure}[htbp]
    \centering
    \includegraphics[width=0.85\linewidth]{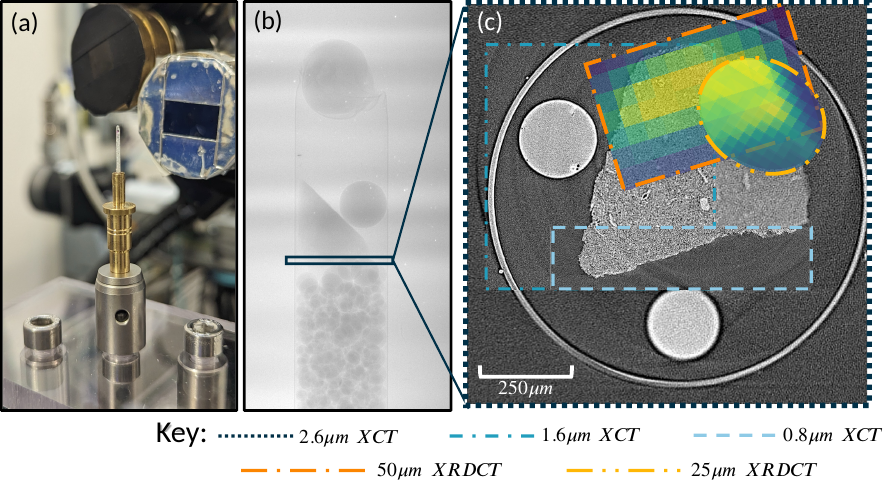}
    \caption{Summary of the datasets collected. (a) shows the experimental setup of a borosilicate capillary containing the Zn-doped Zeolite 13X sample, along with a 0.5mm ruby sphere. (b) is a single projection of the sample capillary from the I13-2 beamline at a 0.8125 micron pixel-size. (c) displays subslices of 3 XCT resolutions (0.8125, 1.625 and 2.6 micron pixel-size) and both XRD-CT diffraction spot-sizes ($25\mu m$ and $50\mu m$) for sodium phase of the sample here. The viridis colormap is used to identify XRD-CT against XCT. The XCT and XRD-CT datasets have been spatially aligned and registered with the help of the fiducial markers.}
    \label{fig:multimodal_fig}
\end{figure}

\FloatBarrier

\section*{Methods}



\subsection*{Zn-13X Sample Preparation}

Molecular sieve 13X (13X) powder from Sigma-Aldrich (Germany) was placed into a pellet die of 5mm diameter. The die was placed into a hydraulic press (Specac Atlas), and exposed to a force of 0.4 tonnes for 30 seconds. The resulting pellet was approximately sized $5mm$ in diameter and $3.5mm$ in length. The Zeolite 13X pellet was placed in a $0.5M$ aqueous solution of zinc chloride ($ZnCl_2$) for 15 minutes, before being dried at $160\degree C$ for 1 hour in an oven. The dried Zn-13X pellet was then cleaved, with a fragment roughly just under $1mm$ in size selected from the edge of the pellet as such that the pellet has a zinc-doped region from reacting with the $Zn^{2+}$ ions in solution, and a sodium-phase from the centre where the zinc ions did not reach in the 15 minute reaction time. The Zn-13X fragment was then placed in a borosilicate glass capillary measuring $1mm$ internal diameter, along with a $0.5mm$ Ruby sphere to act as a fiducial marker, before being filled with NIST standard borosilicate glass beads of diameter $0.1-0.4mm$, sealed with superglue and mounted onto a pinhead. This setup held the sample in the end of the capillary without enveloping the sample, allowing for clearer imaging and diffraction measurements, and could be transferred easily between beamlines. A $1mm$ diameter capillary was chosen to fit within the field-of-view of the imaging camera (pco.edge 5.5) used on the both the I13-2 and DIAD beamlines, at all but one of the magnifications. The borosilicate glass capillary also features a wall thickness of 0.01mm which helps to minimise photon scattering and attenuation from the capillary. 

\subsection*{I13-2: X-ray Computed Tomography}

Multi-resolution XCT imaging was carried out on the I13-2 beamline at Diamond Light Source. I13-2 uses a series of scintillating objective lenses to increase magnification of the sample. Incident X-rays transmit through the sample, hit the scintillating lens and pass through the optical components before illuminating the pco.edge 5.5 imaging camera which sits perpendicular to the incident X-ray beam. Magnification of 1.25x is achieved through the use of a $500\mu m$ thick cerium-doped Lutetium Aluminium Garnet (LuAG) lens; 2x is achieved with a $500 \mu m$ thick LuAG lens; 4x through a $100\mu m$ thick LuAG lens, and 10x through a $25\mu m$ europium-doped Gadolinium Gallium Garnet (GGG:Eu). These magnifications are in addition to a 2x intermediate tube lens, giving final magnifications of 2.5x, 4x, 8x and 20x and pixel-sizes of $2.6\mu m$, $1.625\mu m$, $0.8125\mu m$ and $0.325\mu m$ respectively. The I13-2 beamline is equipped with an undulator insertion device source, which after tuning provides a mean energy of $27$keV. For each tomography scan, 4000 projections were taken as the sample rotated through $0-180\degree$. This number of projections was chosen to satisfy a maximum of one pixel between each angular projection to ensure optimal 3D reconstructions. Exposure time for each scan varied to ensure a roughly consistent amount of detector counts were achieved across all four resolutions and magnifications, and were as follows: 2.5x = 0.15s; 4x = 0.024s; 8x = 0.12s; 20x = 1s.

For each resolution, 4000 projections of dimensions 2560x2160 were acquired, with a rotational angle of $0.045\degree$ between each projection. This volume was then reconstructed using a filter back projection algorithm (FBP), producing a volume of 2510x2510x2110 for each resolution. Reconstructions are performed using the Savu Python package, with full descriptions of the process lists used described in \textit{Usage Notes}.\cite{wadeson_savu_2016} Both the raw data, and reconstructed volumes are available in the \textit{Data Records} section, along with the code used to perform the reconstructions.

\subsection*{DIAD: X-ray Computed Tomography and Diffraction-Tomography}

Both XCT, and XRD-CT were carried out on DIAD (K11). DIAD is a fixed-gap hybrid wiggler sourced beamline where the photon fan is partially split into two parallel beams using two flat and two bending striped mirrors - one beam is used for imaging and the other for diffraction. Both beams are independently configurable with the imaging beam able to do both monochromatic and pink beam. The diffraction beam is then focused again using a pair of Kirkpatrick-Baez (KB) mirrors to be incident on the sample in the same place as the full-field imaging beam, providing quasi-simultaneous imaging and diffraction measurements at the same point in space. \cite{reinhard_beamline_2021} For imaging DIAD uses a pco.edge 5.5 camera, and for its diffraction measurements DIAD uses a PILATUS3 X CdTe 2M detector.

DIAD uses a $10\mu m$ thick terbium-doped Gadolinium Gallium Garnet (GGG:Tb) scintillator for its imaging experiments, and is capable of producing a pixel-size of $0.54\mu m$ and an imaging FOV of $1.4mm$x$1.2mm$. 5000 projections of $2560$x$2160$ pixels with an exposure time of $0.05$s and an angular rotation of $0.036\degree$ between each projection was reconstructed into a volume of $2560$x$2560$x$2160$ voxels using an FBP reconstruction algorithm.

XRD-CT is a technique that allows 3D reconstruction of an object based upon the scattering of incident X-rays as opposed to the absorption of X-rays as in XCT. DIAD is able to perform this technique by scanning the diffraction beam across the imaging FOV, rotating the sample, and re-scanning the diffraction beam across the sample, creating a series of diffraction scans at different angular projections.

Using the KB mirrors, a diffraction spot-size of either $25$x$25\mu m$ or $50$x$50\mu m$ can be achieved within a $1mm$x$1mm$ FOV. At a spot size of $25\mu m$, $40$ diffraction patterns were recorded horizontally across the sample at 80 angular projections between $0-360\degree$ rotation and with an exposure time of 10s per point. For a spot size of $50\mu m$, $20$ diffraction patterns were recorded across the sample at 40 projections between $0-360\degree$ rotation and with an exposure time of 10s. Recording 40 diffraction patterns across 80 projections with a 10s exposure time takes just under 9 hours per scan. For the 1mm capillary used in this experiment, mapping the whole sample at $25\mu m$ spot size would take over two weeks of constant acquisition which is not feasible, whereas mapping a 1mm sample at $50\mu m$ spot size, 20 patterns across 40 projections takes a more reasonable 2 days. All projections for both spot-sizes were acquired throughout an angular range of $0-360\degree$ in order to account for disparate absorption paths of scattered X-rays across the scan path, resulting from the angled position of the diffraction detector.

DIAD uses azimuthal integration to process the raw diffraction data from the detector to a 1D diffraction pattern using the PyFAI Python package.\cite{kieffer_pyfai_2012}  The 1D diffraction pattern is binned into 5000 intervals in $2\theta$ or $q$. For the $25\mu m$ spot size, the reduced XRD-CT dataset has a size of 5000x40x80, and for the $50\mu m$ spot size was 5000x20x40.

\begin{figure}[htbp]
    \centering
    \includegraphics[width=0.75\linewidth]{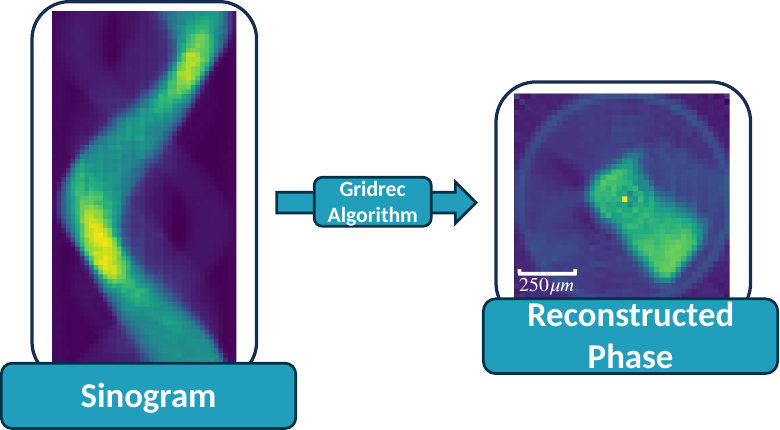}
    \caption{Example of a sinogram at q=1.649 for a spot size of $25\mu m$ and the reconstructed image using a gridrec reconstruction algorithm in the TomoPy Python package. The reconstructed image shows the Sodium phase in that $25\mu m$ region of the Zn-doped Zeolite 13X sample. The data used to create this figure file number \textit{43322} in \href{https://zenodo.org/records/13329639}{10.5281/zenodo.13329639}.\cite{green_xrdct_diad}}
    \label{fig:diad_sino_recon}
\end{figure}

Plotting the intensity of the 1D diffraction signal for a given scattering angle for each of the 40 points scanned against the sample rotation produces a sinogram as shown in Figure \ref{fig:diad_sino_recon}. These sinograms can be reconstructed into an image using a gridrec algorithm in the TomoPy Python package.\cite{gursoy_tomopy_2014, marone_regridding_2012}.

\FloatBarrier

By identifying which peaks in the diffraction pattern correspond to which phase, specific phase-based reconstructions can be performed by using a gridrec algorithm on the average of the sinograms that correspond to that peak in q-space. Additional powder diffraction data on pure Zn-13X and pure 13X was obtained to identify the main peaks in $q$-space in the diffraction data for both the Zn-phase and Na-phase as described in the \textit{Technical Validation}. Using these q-values, the sinograms around these peaks were averaged and reconstructed using a gridrec algorithm, producing distinct zinc and sodium-based phases within the Zn-13X sample used. Figure \ref{fig:diad_sino_recon} shows distinct sinograms for the Zn and Na-phases within the sample which identify distinct Zn and Na-phases within one of the $25\mu m$ resolution region of interest (ROI) datasets.

\section*{Data Records}



XCT data from I13-2 at the four different resolutions can be found as a series of Zenodo entries, as described in Table \ref{tab:i13_zenodo} below. Each raw entry contains the code executed using the savu reconstruction software and the associated savu process list to reproduce the reconstructed data. Raw data exists as a .hdf file, and the reconstructed data is stored as a .h5 - both also have an associated .nxs file.

\begin{table}[htbp]
    \centering
    \begin{tabular}{c|c|c|c|c}
        Magnification & Pixel-Size ($\mu m)$ & Field-of-view (mm) & Zenodo DOI (Raw) & Zenodo DOI (Processed) \\
        \hline
         2.5x & 2.6 & $6.7 \times 5.6$ & \href{https://zenodo.org/records/13327891}{10.5281/zenodo.13327891}\cite{green_xct_i13-2_26_raw} & \href{https://zenodo.org/records/12206815}{10.5281/zenodo.12206815}\cite{green_xct_i13-2_26} \\
         4x  & 1.625 & $4.2 \times 3.5$ & \href{https://zenodo.org/records/13327931}{10.5281/zenodo.13327931}\cite{green_xct_i13-2_1625_raw} & \href{https://zenodo.org/records/13327651}{10.5281/zenodo.13327651}\cite{green_xct_i13-2_1625} \\
         8x & 0.8125 & $2.1 \times 1.8$ & \href{https://zenodo.org/records/13327958}{10.5281/zenodo.13327958}\cite{green_xct_i13-2_08125_raw} & \href{https://zenodo.org/records/13327682}{10.5281/zenodo.13327682}\cite{green_xct_i13-2_08125} \\
         20x & 0.325 & $0.83 \times 0.7$ & \href{https://zenodo.org/records/13327961}{10.5281/zenodo.13327961}\cite{green_xct_i13-2_0325_raw} & \href{https://zenodo.org/records/13327692}{10.5281/zenodo.13327692}\cite{green_xct_i13-2_0325} \\
    \end{tabular}
    \caption{Summary of the four different XCT entries on Zenodo from I13-2 beamline, including the magnification, pixel-size, field-of-view, and Zenodo entries for the raw and processed data}
    \label{tab:i13_zenodo}
\end{table}

Figure \ref{fig:I13-2_XCT_all} below displays the corresponding subslice of all four I13-2 pixel-sizes acquired after being spatially aligned.

\begin{figure}[htbp]
    \centering
    \includegraphics[width=0.95\linewidth]{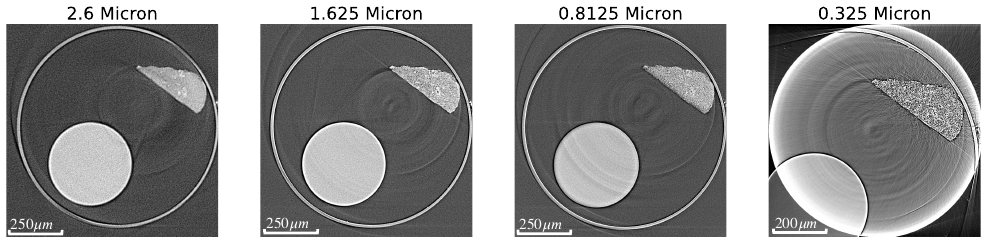}
    \caption{Side-by-side images showing the same cross-section of the sample across all four different resolution on I13-2. Note that the smallest pixel-size of 0.325 microns a FOV smaller than the size of the sample and results in some exclusion of the sample.}
    \label{fig:I13-2_XCT_all}
\end{figure}

\FloatBarrier

Processed XRD-CT data for both the Zn and Na phases at a spot size of 25 and 50 microns are available at \href{https://zenodo.org/records/13329639}{10.5281/zenodo.13329639} as a series of .nxs files.\cite{green_xrdct_diad} The sinograms pre-reconstruction are also included in the same repository as .nxs files. Powder X-ray diffraction data from a Rigaku SmartLab Diffractometer can be found at \href{https://zenodo.org/records/13329670}{10.5281/zenodo.13329670} as a series of .asc files to assist with performing the phase-based reconstructions of the XRD-CT data along with a Jupyter Notebook to reproduce the analysis.\cite{green_zn13x_xrd}

Reconstructed XCT data from the DIAD beamline can be found at \href{https://zenodo.org/records/13349402}{10.5281/zenodo.13327988} as a .h5 and .nxs file, whilst the raw data is available at \href{https://zenodo.org/records/13329625}{10.5281/zenodo.13329625} as a .h5 file.\cite{green_xct_diad_054, green_xct_diad_054_raw}

\section*{Technical Validation}


\subsection*{Zn-13X Characterisation}

To confirm the existence of zinc in the partially Zn-doped 13X sample used, powder XRD measurements were obtained on both pure-13X, and pure-Zn-13X using a Rigaku SmartLab Diffractometer using a copper X-ray source of wavelength $1.5406$\r{A}. Pure Zn-13X was obtained by following the experimental procedure described in \textit{Methods}, except 13X powder was left for 24 hours instead of 15 minutes to allow for more complete ion-exchange.\cite{david_cr13x_2022} All the powder XRD data is available as Zenodo entry at \href{https://zenodo.org/records/13329670}{10.5281/zenodo.13329670}. The powder diffraction showed a shift in the peaks between the Zn-doped 13X and pure 13X, with the peaks in the Zn-13X powder diffraction pattern moving to a slightly higher q-value than those from the pure 13X powder diffraction pattern, validating the presence of two distinct phases. The magnitude of the shift in q is $q=0.01$.

\subsection*{XCT Validation}

The I13-2 beamline has a synchrotron undulator X-ray source that is situated approximately 230m from the sample position. The X-ray beam from the undulator source is reflected by a Pt X-ray mirror which directs the beam to the experimental station. At first, a 1mm thick knife edge (JJ X-ray) is used for focusing the objectives. A series of images of the knife edge are collected at different focusing distances (steps of 0.05mm) on either side of the visually estimated value. This is repeated without the knife edge (flat-field images) and then without the beam (dark-field images). Once the flat and dark field correction is performed on the images, a line profile across the knife edge for different focusing position is plotted. The focusing position corresponding to the image with sharpest drop in the intensity is selected as the focusing position for the objective. This is repeated for the other objectives. To correct for the barrel and pin cushion distortion of the objectives, images are collected with a grid-like dot pattern having a period (of the dot) varying from $100-25\mu m$ and dot diameter of $10-3\mu m$.  Using these images distortion correction parameters are calculated and later integrated into the reconstruction pipeline.\cite{vo_radial_2015}

Using a standard sample made up of a $300\mu m$ metal ball in a capillary, a tomography acquisition was carried out. By tracking the centroid of this ball at different angles of the tomography, the rotation axis tilts along the beam and across the beam are measured and corrected. This sets the rotation axis parallel to the detector column and perpendicular to the detector rows.

A similar method is used on DIAD to focus the sample and align the imaging camera with the centre of rotation.

\subsection*{DIAD Diffraction-Tomography Validation}

The capability of DIAD to quasi-simultaneously acquire imaging and diffraction measurements means the imaging camera lies in-line with the incident X-ray beam, and the diffraction detector lies away at an angle on an industrial robot arm. Consequently, the centre of the diffraction beam is not in-line with the detector, allowing only partial Debye-Scherrer rings to be collected. \cite{reinhard_beamline_2021} The diffraction detector is held on an industrial robot arm which has been found to introduce small spatial movements over the acquisition period, but these are an order of magnitude smaller than the diffraction detector pixel size. \cite{reinhard_flexible_2022} DIAD scans the diffraction beam across the imaging field-of-view, which means the beam-detector geometry is not fixed. Consequently each diffraction beam point across imaging FOV must be calibrated as a separate sample-detector geometry to perform successful azimuthal integration in the data reduction process.

\section*{Usage Notes}


Raw and processed XCT data is provided as a series of Hierarchical Data Format (.h5 and .hdf files) and NeXus (.nxs) files. The recommended software for visualising the raw data and processed data as .h5/.hdf5 files is the open-source ImageJ/Fiji package (\hyperlink{https://imagej.net/software/fiji/}{https://imagej.net/software/fiji/}). The Data Analysis WorkbeNch, DAWN (\hyperlink{https://dawnsci.org}{https://dawnsci.org}) is best for loading in NeXus (.nxs) files. Savu scripts and process lists are provided to perform the reconstructions using savu alongside raw data in the Zenodo entries. Sinograms from the DIAD XRD-CT are stored as .nxs files and can be viewed in DAWN. A Jupyter Notebook is provided that describes the loading of the raw data and reconstruction process using TomoPy. Reconstructions can be viewed as .nxs files in DAWN. A Jupyter notebook is also provided that shows the process used to align the different resolutions of XCT acquired on I13-2, and the normalisation used to ensure roughly consistent colours across all the reconstructed 2D slices used to create Figure \ref{fig:multimodal_fig}. Powder diffraction data is provided as a series of ASCII (.asc) files for pure 13X, pure Zn-13X, Zn-13X sample, and the empty capillary. An example Jupyter Notebook is provided in the Zenodo entry to load and visualise the data.

\begin{figure}[htbp]
\centering
\begin{subfigure}[b]{0.9\textwidth}
   \centering
   \caption{XCT Pipeline}
   \includegraphics[width=0.6\linewidth]{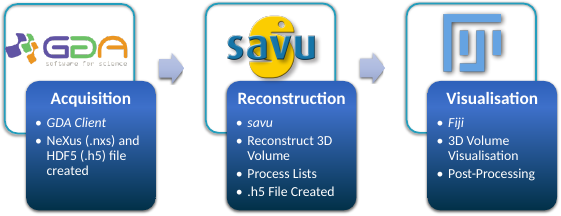}
   \label{fig:xct_pipeline} 
\end{subfigure}

\begin{subfigure}[b]{0.9\textwidth}
   \centering
   \caption{XRD-CT Pipeline}
   \includegraphics[width=0.8\linewidth]{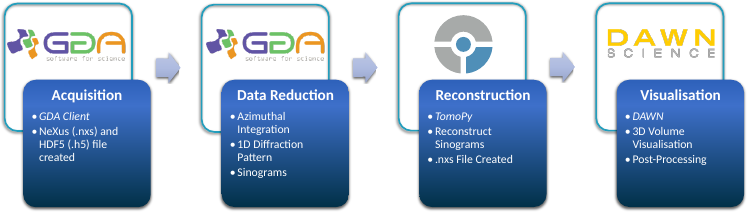}
   \label{fig:difftomo_pipeline}
\end{subfigure}

\caption{Data processing pipelines for both (a) XCT data and (b) XRD-CT data from acquisition through to reconstruction to post-processing. Softwares and file formats are also indicated}
\label{fig:pipelines}
\end{figure}

\FloatBarrier

Figure \ref{fig:pipelines} above summarises the different software used and visualises the pipelines that are used to acquire, process and visualise the XCT and XRD-CT data. The pipelines are specific to dataset modalities and are the same across different beamlines.

\section*{Code availability}

Code to reproduce reconstructed data from raw data, and to visualise all the data can be found within the relevant Zenodo entries and on the associated \href{https://github.com/calum-green/XCT-XDRCT_paper_code}{GitHub page} that houses a duplicate copy of the code associated with each Zenodo entry in one place.


\bibliography{references}


\section*{Acknowledgements}

This work was part funded by the EPSRC prosperity partnership with Imperial College, INFUSE, Interface with the Future - Underpinning Science to Support the Energy transition EP/V038044/1 under CG's studentship STU0446. The authors would like to acknowledge the Diamond Light Source for access to the Dual Imaging And Diffraction (DIAD) K11 and I13-2 beamlines under proposal MG32980. The authors would also like to thank Dr Daniel Nye for help on the Rigaku SmartLab Diffractometer in the Materials Characterisation Laboratory at the ISIS Neutron and Muon Source, and Joel Lees-Massey for assistance with reconstructing the DIAD XRD-CT work. This work was partially funded by the Rutherford Appleton Laboratory and Faraday Institution, through JLH's emerging leader fellowship FIELF001. DD would also like to acknowledge the support received from the Royal Academy of Engineering via his Research Chair in Complex Engineering Interfaces. This work was part funded by the National Research Facility for Lab X-ray CT (NXCT), EP/T02593X/1.

\section*{Author contributions statement}

CG, JLH, and DD conceived the original experimental idea, with CG and JLH submitting proposal MG32980 and conducting the two experiments. SA, LP, and AL assisted the DIAD experiment, and SM assisted the I13-2 experiment. CG and JLH conducted post-processing of the raw data from both experiments. CG wrote the initial draft, and all authors reviewed and contributed to the manuscript.

\section*{Competing interests}

The authors declare no competing interests.


\end{document}